\documentclass[letterpaper, 10 pt, conference]{ieeeconf}  

\IEEEoverridecommandlockouts                              

\usepackage{multicol}
\usepackage[bookmarks=true]{hyperref}
\usepackage{subfigure}
\usepackage{graphicx}
\usepackage{xcolor}
\usepackage{mathtools}
\usepackage{caption}
\usepackage{gensymb}
\usepackage{algorithm}
\usepackage{amssymb}
\usepackage{amsmath}
\usepackage{algpseudocode}
\usepackage[keeplastbox]{flushend}
\algdef{SE}[SUBALG]{Indent}{EndIndent}{}{\algorithmicend\ }%
\algtext*{Indent}
\algtext*{EndIndent}

\title{\LARGE \bf
Learning to Guide Human Attention on Mobile Telepresence \\Robots with $360\degree$ Vision
}

\author{Kishan Chandan, Jack Albertson, Xiaohan Zhang, Xiaoyang Zhang, Yao Liu, Shiqi Zhang
\thanks{All authors are with the State University of New York (SUNY) at Binghamton, Binghamton, NY 13902 USA}
\thanks{  Email: {\tt\small \{kchanda2;jalbert5;xzhan244;xzhan211; yaoliu;zhangs\}@binghamton.edu} }%
}

\begin{document}

\maketitle
\thispagestyle{empty}
\pagestyle{empty}

\begin{abstract}
Mobile telepresence robots (MTRs) allow people
to navigate and interact with a remote environment that is in a place other than the person's true location. 
Thanks to the recent advances in $360\degree$ vision, many MTRs are now equipped with an all-degree visual perception capability. 
However, people's visual field horizontally spans only about $120\degree$ of the visual field captured by the robot. 
To bridge this observability gap toward human-MTR shared autonomy, we have developed a framework, called GHAL360, to enable the MTR to learn a goal-oriented policy from reinforcements 
for guiding human attention using visual indicators. 
Three telepresence environments were constructed using datasets that are extracted from Matterport3D and collected from a real robot respectively.
Experimental results show that GHAL360 outperformed the baselines from the literature in the efficiency of a human-MTR team completing target search tasks. A demo video is available: {\color{blue} \texttt{https://youtu.be/aGbTxCGJSDM}}

\end{abstract}

\section{Introduction}


Telepresence is an illusion of spatial presence, at a place other than the true location~\cite{minsky1980telepresence}.
Mobile telepresence robots~(\textbf{MTRs}) enable a human operator to extend their perception capabilities along with the ability of moving and actuating in a remote environment~\cite{kristoffersson2013review}.
The rich literature of mobile telepresence robotics has demonstrated applications in domains such as offices~\cite{desai2011essential,takayama2012mixing}, academic conferences~\cite{neustaedter2016beam,rae2017robotic}, elderly care~\cite{cesta2010enabling,sabelli2011conversational}, and education~\cite{ahumada2019going,rae2015framework}.

Recently, researchers have equipped MTRs with a $360\degree$ camera to perceive the entire sphere of a remote environment~\cite{zhang2018360,hansen2018head}.
In comparison to traditional monocular cameras (including the pan-tilt ones), $360\degree$ visual sensors have equipped the MTRs with the capability of omnidirectional visual scene analysis. 
However, the human vision system, by nature, is not developed for, and hence not good at processing $360\degree$ visual information. 
For instance, computer vision algorithms can be readily applied to visual inputs with varying spans of degrees, whereas the binocular visual field of the human eye spans only about $120\degree$ of arc~\cite{ruch1960medical}. 
The portion of the field that is effective to complex visual processing is even more limited, e.g., only $7\degree$ of visual angle for facial information~\cite{papinutto2017facespan}. 
\emph{How can one leverage the MTRs' $360\degree$ visual analysis capability to improve the human perception of the remote environment?}
One straightforward idea is to project $360\degree$ views onto equirectangular video frames (e.g., panorama frames).
However, people tend to focus on small portions of equirectangular videos, rather than the full frame~\cite{gaddam2016tiling}. 
This can be seen in Google Street View, which provides only a relatively small field of view to the users, even though the $360\degree$ frames are readily available. 
Toward long-term shared autonomy, we aim to bridge the \textbf{observability gap} in human-MTR systems equipped with $360\degree$ vision. 



In this paper, we propose a framework, called Guiding Human Attention with Learning in $360\degree$ vision (\textbf{GHAL360}). 
To the best of our knowledge, GHAL360, for the first time, equips MTRs with simultaneous capabilities of $360\degree$ scene analysis and guiding human attention. 
We use an off-policy reinforcement learning (RL) method~\cite{sutton2018reinforcement} to compute a policy for guiding human attention to areas of interest. 
The policies are learned toward enabling a human to efficiently and accurately locate a target object in a remote environment. 
More specifically, $360\degree$ visual scene analysis produces a set of detected objects, and their relative orientations; and the learned policies map the current world state, including both human state (current attention), and the scene analysis output, to an action of visual indication. 

We have evaluated GHAL360 using target search tasks. 
Three virtual environments have been constructed for demonstration and evaluation purposes---two of them using the Matterport3D datasets \cite{Matterport3D}, and one using a dataset collected from a mobile robot platform.
We have compared GHAL360 with existing methods including~\cite{kaplan1997internet,heshmat2018geocaching}. 
From the results, we see that GHAL360 significantly improved the efficiency of the human-MTR system in locating target objects in a remote environment. 

\begin{figure*}[tb]
    \begin{center}
    \includegraphics[width=16cm]{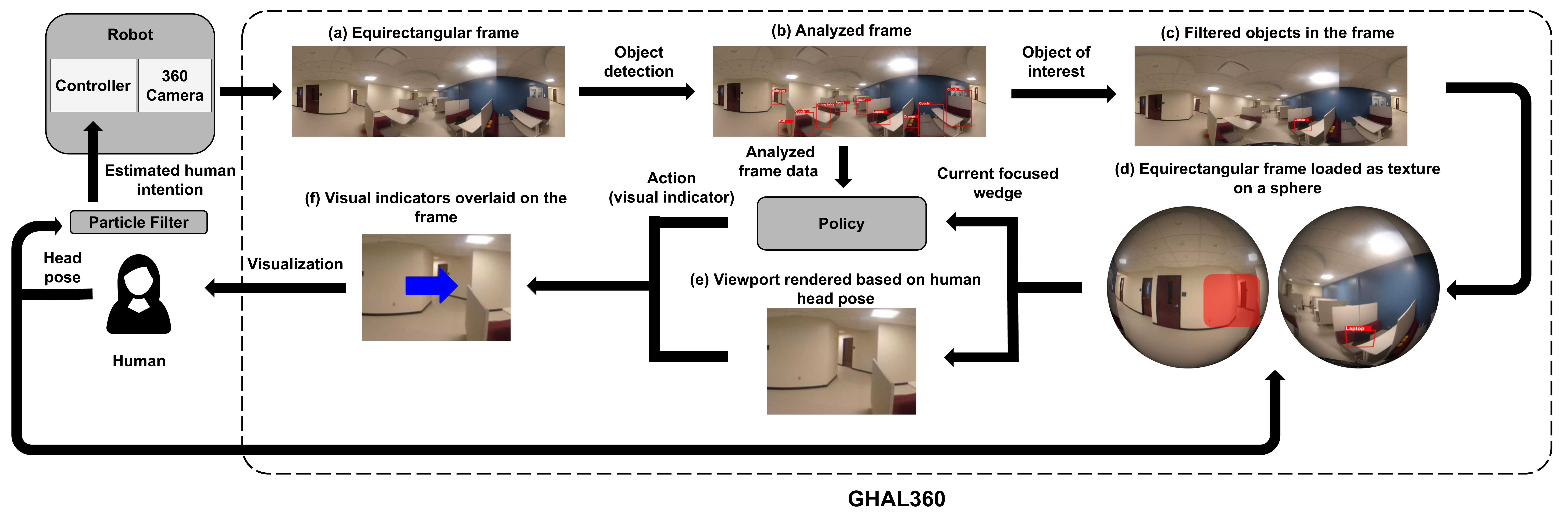}
    \caption{
The input of GHAL360 includes live 360$\degree$ video frames of the remote environment from MTR, and human head pose. 
(a) \textbf{Equirectangular frames} encode the 360$\degree$ information using equirectangular projection. 
(b) Those frames are analyzed using an object detection system. 
(c) The detected objects are then filtered to highlight only the objects of interest, which is a task-dependent process. 
(d) After the human's display device receives the equirectangular frames, GHAL360 constructs a sphere, and uses the frames as the texture of the sphere. 
A human may only view a portion of the sphere, with a limited field of view, e.g., as indicated in the red-shaded portion. 
(e) The portion viewed by the human is called a ``\textbf{viewport}'', which is rendered based on the head pose of the human at runtime. 
(f) GHAL360 overlays visual indicators in real time to guide the human attention toward the object of interest while providing an immersive 360$\degree$ view of the remote environment.     
    }
    \label{fig:scene_visualizer_360}
    \end{center}
    \vspace{-1.9em}
\end{figure*}

\section{Related Work}

There is rich literature on the research of mobile telepresence robots (MTRs). 
Early systems include a telepresence system using a miniature car model~\cite{kaplan1997internet}, and a telepresence robot for enabling remote mentoring, called Telementoring~\cite{agarwal2007roboconsultant}.
Those systems used monocular cameras that produce a narrow field of view over remote environments. 
MTR systems were developed to use pan-tilt cameras to perceive the remote environment~\cite{paulos2001social,baker2004improved,kuzuoka2000gestureman}, where the human operator needs to control both the pan-tilt camera and the robot platform.
To relieve the operator from the manual control of the camera, researchers leveraged head motion to automatically control the motion of pan-tilt cameras using head-mounted displays~\cite{bolas1990head,aykut2018delay}.
While such systems eliminated the need of controlling a pan-tilt camera's pose, they still suffered from the limited visual field. In comparison to the above-mentioned methods and systems, GHAL360 (ours) leverages $360\degree$ vision to enable the MTR to perceive the entire sphere of the environment. 

To increase a robot's field of view, researchers have equipped mobile robots with multiple cameras~\cite{ikeda2003high,karimi2018mavi}, and wide-angle cameras~\cite{10.1145/238386.238402,mair1997telepresence}.
Recently, researchers have developed MTRs with $360\degree$ vision systems~\cite{heshmat2018geocaching,hansen2018head}.
Examples include the MTRs with $360\degree$ vision for redirected walking~\cite{zhang2018360}, and virtual tours~\cite{oh2018360}. 
In comparison to their systems, GHAL360 equips the MTRs with a capability of $360\degree$ scene analysis, and further enables the robot to learn to use the outputs of scene analysis to guide human attention. 
As a result, GHAL360 performs better than those methods in providing the human with more situational awareness. 
It should be noted that transmitting $360\degree$ visual data brings computational burden and bandwidth requirement to the robot~\cite{zhou2020adap,hosseini2016adaptive}. 
Such system challenges are beyond the scope of this work.

Rich literature exists regarding active vision within the computer vision and robotics communities~\cite{aloimonos1988active, andreopoulos201350, dickinson1997active}, where active vision methods enable an agent to plan actions to acquire data from different viewpoints. 
For instance, recent research has showcased that an end-to-end learning approach enables an agent to learn motion policies for active visual recognition~\cite{jayaraman2018end}. 
Work closest to GHAL360 is an approach for automatically controlling a virtual camera to choose and display the portions of video inside the $360\degree$ video~\cite{su2016pano2vid, su2017making}.
Compared with those methods on active vision, GHAL360 has a human in the loop, i.e., it takes the current human state into account when using visual indicators to guide the human's attention. 
Another difference is that GHAL360 is goal-oriented, meaning that its strategy of guiding human attention is learned toward achieving a specific goal (in our case, the goal is to locate a specific target object).

\section{Framework and System}
In this section, we present our framework called GHAL360, short for Guiding Human Attention with Learning in $360\degree$ vision, that leverages the MTR's 360$\degree$ scene analysis capability to guide human attention to the remote areas with potentially rich visual information. 
Fig.~\ref{fig:scene_visualizer_360} shows an overview of GHAL360.

\subsection{The GHAL360 Framework}

Algorithm~\ref{alg:visual_indicator_algorithm} delineates the procedure of GHAL360 and explains the different stages as well as the data flow at each stage.
The input of GHAL360 includes $\kappa$, the name of a target object of interest as a string, and $objDet$, a real-time object detection system.

After a few variable initializations (Lines~\ref{line:initialize_C}-\ref{line:initialize_P}), GHAL360 enters the main control loop in Line~\ref{line:main_while}.
The main control loop continues until the human finds the target object in the remote environment.
Once a new frame ($\mathcal{F}$) is obtained, GHAL360 analyzes the remote scene using an object detection system and stores the objects' names and their locations in a dictionary $\mathcal{P}$ (Line~\ref{line:objdet}).
The location of $\kappa$ is then stored in $\mathcal{L}$ and is used to draw the bounding box over $\mathcal{F}$.

If the target object is detected in the current frame (Line~\ref{line:object}), Lines~\ref{line:get_state}-\ref{line:policy_update} are activated for guiding human attention.
In Line~\ref{line:get_state} the current egocentric state representation ($s$) of the world  is obtained using the $getState$ function.
Based on $s$, the policy generated from the RL approach ($\pi$) returns an action, which is a guidance indicator to be overlayed on the current viewport ($\mathcal{F^{'}}$), and then $F^{'}$ is presented to the user via the interface.
After executing every action $a$, GHAL360 updates the Q-values for the state-action pair using the observed reward ($r$) and the next state ($s'$) in Line~\ref{line:update_q_value}.
Finally, using the new Q-values, the policy ($\pi$) is updated, and in the next iteration, an optimal action is selected using the updated policy.
The human head orientation ($\mathcal{H}$) is sent to the particle filter as evidence to predict the human motion intention ($\mathcal{I}$).
The controller then returns the control signal as $\boldsymbol{\mathcal{C}}$ based on $\mathcal{I}$ to control the robot motion in the remote environment.


\begin{algorithm}[tb] \footnotesize
\caption{}\label{alg:visual_indicator_algorithm}
\begin{algorithmic}[1]
        \Require {$\kappa$, $objDet$ }
        \State{Initialize an empty list $\boldsymbol{\mathcal{C}}$ to store the control commands as $\emptyset$}\label{line:initialize_C}
        \State{Initialize a quaternion $\mathcal{H}$ (Human Pose) as (0,0,0,0)}
        \State{Initialize a 2D image $\mathcal{F}$ with all pixel values set as 0}
        \State{Initialize a dictionary $\mathcal{P} =$\{\}}\label{line:initialize_P}
        \State{$\pi \xleftarrow{}$ random policy}\Comment{Initialize a policy}
        
        \While{True}\label{line:main_while}

            \If{new frame is available}\label{line:if_frame}
                \State{Obtain current frame $\mathcal{F}$ of the remote environment}\label{line:getFrame}
                \State{$\mathcal{P} \xleftarrow{} objDet(\mathcal{F})$ }\label{line:objdet}\algorithmiccomment{Section~\ref{sec:scene_analysis}}
                \State{$\mathcal{L}$ $=$ $\mathcal{P}[\kappa]$}\label{line:get_obj_interest} \algorithmiccomment{Get the location of the object of interest}
                \State{Overlay bounding box for object of interest over $\mathcal{F}$}\label{line:overlay_bounding_box} 
                \State{Render the frame as a sphere}\Comment{Section~\ref{sec:equi_rendering}}\label{line:spherical_format}
                \State{Obtain current human head pose $\mathcal{H}$}\label{line:getHeadPose}
                         \State{Render a \emph{viewport} $\mathcal{F}^{'}$ to match the human operator's head pose}\label{line:crop_frame}
                 \If{$\kappa$ in $\mathcal{P}$} \label{line:object}
                \State{$s$ = $getState(\mathcal{P}, \kappa,\mathcal{H})$} \Comment{Egocentric state representation} \label{line:get_state}
                
                \State{$a = \pi(s)$} \Comment{Section~\ref{subsection:rl_policy}}\label{line:rl_action}
                \If{$a$ == $left$ or $right$}
                    \State{Overlay the visual indicator on $\mathcal{F}^{'}$}
                \EndIf
                \State{Present $\mathcal{F}^{'}$ to the user via the interface}\label{line:return_left}
                \State{Observe $r$ and $s'$} \Comment{Immediate reward and next state}
                \State{$Q(s, a) \leftarrow Q(s, a) + \alpha \left[r + \gamma \max_{a'} Q(s', a') - Q(s, a) \right]$}\label{line:update_q_value}
                \State{$\pi(s) \xleftarrow{} argmax_{a}Q^{*}(s,a)$} \Comment{Update policy}\label{line:policy_update}
                  \EndIf

                \State{$\mathcal{I} \xleftarrow{} pf(\mathcal{H})$} \Comment{Section~\ref{sec:particle_filter}}\label{line:particle_filter}
                \State{$\boldsymbol{\mathcal{C}} \xleftarrow{} $controller($\mathcal{I}$)} \Comment{Control signal based on human intention}\label{line:controller}
                \For {\textbf{each} $c$ $\in$
                $\boldsymbol{\mathcal{C}}$}\label{line:for_each_cntrlcmd}\Comment{Section~\ref{sec:telepresence_interface}}
                    \State{$\psi \xleftarrow{} \tau(c)$}\label{line:teleopmsg}\algorithmiccomment{Generate a teleop message}
                    \State{$\psi$ is sent to the robot for execution}
                    \State{Updated map is presented via the interface}
                \EndFor
             \EndIf

        \EndWhile
        
\end{algorithmic}
\end{algorithm}

In the following subsections, we describe the key components of GHAL360.

\subsection{Scene Analysis}\label{sec:scene_analysis}
Once a new frame ($\mathcal{F}$) is obtained, GHAL360 analyzes the remote scene information using an object detection system, where we use a state-of-the-art convolutional neural network (CNN)-based framework \textbf{YOLOv3}~\cite{yolov3}.
All the objects (keys) in the frame along with their locations (values) are stored in a dictionary $\mathcal{P}$ (Line~\ref{line:objdet}).
In our implementation, we divide the $360\degree$ frame into eight equal $45\degree$ wedges.
These eight wedges together represent all the four cardinal and four intercardinal directions, i.e. [N, S, E, W, NE, NW, SE, SW].
Every wedge can have four different values based on the objects detected in that wedge:
$$
    W:  \{w_{0},w_{1},w_{2},w_{3}\}
$$
\begin{itemize}
    \item $w_{0}$: represents a wedge with no objects detected
    \item $w_{1}$: represents a wedge with clutter
    \item $w_{2}$: represents a wedge that contains only the object of interest
    \item $w_{3}$: represents a wedge containing clutter as well as the object of interest
\end{itemize}
where clutter indicates that the wedge contains one or more objects that are not the target object.
The output of scene analysis is stored as a vector where each element represents the status of one of the wedge.
The human operator can be focusing on one of these eight wedges.
The $getState$ function in Line~\ref{line:get_state} takes the output of the scene analysis along with the target object and current human head pose as input.
The output of $getState$ function is
an egocentric state representation of the vector representing the locations of objects with respect to the wedge human is currently focusing on ($W_{0}$). 
The generated egocentric version of the vector is then sent to the RL agent as the state representation of the current world, including both the output of scene analysis and the current status of the human operator.

\subsection{Equirectangular Frame Rendering}\label{sec:equi_rendering}



One of the ways to represent 360-degree videos is to project them onto a spherical surface.
Consider the human head as a virtual camera inside the sphere.
Such spherical projection helps to track the human head pose inside the $360\degree$ frames.
In our implementation, we construct a sphere and use $\mathcal{F}$ as the texture information of the sphere.
Inside the sphere, based on the yaw and pitch of the human head, the pixels in the human field of view are identified.
Based on the human operator's head pose obtained in Line~\ref{line:getHeadPose}, the viewport is rendered from the sphere.
We use \textbf{A-Frame}, an open-source web framework for building virtual reality experiences~\footnote{\url{https://aframe.io/}}, for rendering the equirectangular frames to create an immersive $360\degree$ view of the remote environment.

\subsection{Policy for Guiding Human Attention}\label{subsection:rl_policy}



We use a reinforcement learning approach~\cite{sutton2018reinforcement}, Q-Learning, to let the agent learn a policy (for guiding human attention) by interacting with the environment.
The mathematical representation of the state space is: 
$$
    S: W_0 \times W_1 \times \cdots \times W_{N-1}
$$
where $W_{i}$ represents the state of a particular wedge, as defined in Section~\ref{sec:scene_analysis}.
In our case, $N=8$, so $|S|=4^{8}$, and $s \in S$. 
The state space ($S$) of the RL agent consists of a set of all possible state representations of the $360\degree$ frames.
All the state representations are egocentric, meaning that they represent the state of the world~(locations of objects) with respect to the wedge human is currently focusing on. 
The domain consists of three different actions:
$$
A: \{left, right, confirm\}
$$
where $left$ and $right$ actions are the guidance indicators, while the $confirm$ action checks if the object of interest is present in the currently focused wedge or not.
Based on $s$, the policy generated from the RL approach ($\pi$) returns an action ($a$), which can be a guidance indicator to enable the human to find the object of interest in the remote environment.
Once the agent performs the action, it observes the immediate reward ($r$) and the next state ($s'$).
Using the tuple ($s, a, r, s'$), the agent updates the Q-value for the state-action pair.
Finally, the agent updates the policy at runtime based on its interaction with the environment.
We use the BURLAP library~\footnote{\url{http://burlap.cs.brown.edu/}} for the implementation of our reinforcement learning agent and the domain.

\subsection{Particle Filter for Human Intent Estimation}\label{sec:particle_filter}

We use a particle filter-based state estimator to identify the motion intention of the human operator.
In GHAL360, the particle filter is first initialized with $M$ particles:
$$
X = \{x^{0}, x^{1}, \cdots, x^{M-1}\}
$$
where each particle represents a state, which in our case is one of the eight wedges.
Each particle has an associated weight ($\omega^{i}$) and is initialized as $\omega^{i} $=$ \frac{1}{M}$.
The belief state in particle filter is represented as:
$$
B(X_{t}) = P(X_{t} | e_{1:t})
$$
where $t$ is the time step, $e$ is the evidence or the observation of the particle filter, and the belief state $B(X_{t})$ is updated based on the evidence $e_{1}, \cdots, e_{t}$ obtained till time step $t$.
In our implementation, the evidence to the particle filter is the current change in human head orientation (left or right), and the current focused wedge.
As the human starts to look around in the remote environment, based on the evidence the belief is updated:
$$
B^{'}(X_{t+1}) = \sum_{x_{t}} P(X^{'}|x_{t}) B(x_{t}))
$$
After every observation, once the belief is updated, we calculate the density of particles for all the possible states, and the state with more than a threshold percentage of particles (70\% in our case) is considered as the predicted human intention.
Once the particle filter outputs the predicted human intention~(Line~\ref{line:particle_filter}), we use a controller that outputs a control signal to drive the robot base toward the intended area based on the predicted human intention.
In case that the particle density does not reach the threshold in any wedge, the robot stays still until the particle filter can estimate the human motion intention.

\subsection{Telepresence Interface}\label{sec:telepresence_interface}

\begin{figure}[t]
    \begin{center}
    \vspace{.5em}
    \includegraphics[width=8.5cm]{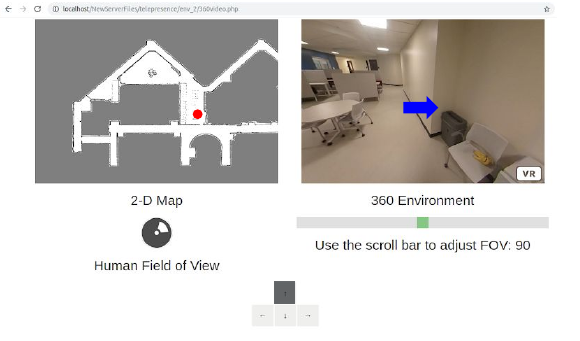}
    \end{center}
    \vspace{-1.5em}
    \caption{\emph{Telepresence Interface} showing the 2D map, the frame of the remote environment, a 2D icon to show human head orientation, a scrollbar to adjust field of view, and a set of icons to indicate the current teleoperation commands given by the human operator.}
    \vspace{-1em}
    \label{fig:interface}
\end{figure}

Based on the action suggested by the policy, GHAL360 overlays visual indicators over the current viewport. This results in a new frame $\mathcal{F^{'}}$ which is presented to the user via the interface~(Fig.~\ref{fig:interface}).
The operator can use ``click'' and ``drag'' operations to look around in the remote environment.
The telepresence interface provides a scroll bar to adjust the field of view of the human operator.
Additionally, the users can easily change the web-based visualization to an immersive virtual reality $360\degree$ experience via head-mounted displays (HMDs) via a \textbf{VR} button at the bottom right of the interface.
In the VR mode, the users can use their head motions to adjust the view of the remote environment.

The telepresence interface also shows the 2D map of the remote environment.
Based on the robot pose, a red circle is overlaid on the 2D map to indicate the live robot location in the remote environment.
Additionally, we show a 2D icon to convey the head orientation of the human operator relative to the map.
The 2D icon also shows the part of the $360\degree$ frame which is in the field of view of the human operator (Fig.~\ref{fig:interface}).
The controller in Line~\ref{line:controller} converts the estimated human intention into control command.
The control commands are converted to a ROS ``teleop'' message and are passed on to the robot to remotely actuate it.
Based on the control command, the robot is remotely actuated and the 2D map is updated accordingly to show the new robot pose.




\section{Experiments}\label{sec:experiments}

We conducted experiments to evaluate GHAL360 in a target-search scenario where a human was assigned the task of finding a target object in the remote environment using a MTR.
We designed two types of simulation environments, 1. \emph{Abstract simulation}, and 2. \emph{Realistic simulation}.
To facilitate learning, we designed \emph{Abstract simulation} to enable the agent to learn a goal-oriented policy for guiding human attention.
The learned system is then evaluated in \emph{Realistic simulation}.
The main difference between \emph{Abstract simulation} and \emph{Realistic simulation} is that the \emph{Abstract simulation} eliminates the real-time visualization to speed up the learning process.

\begin{figure}
\begin{center}
    \subfigure[]
    {\includegraphics[height=2cm]{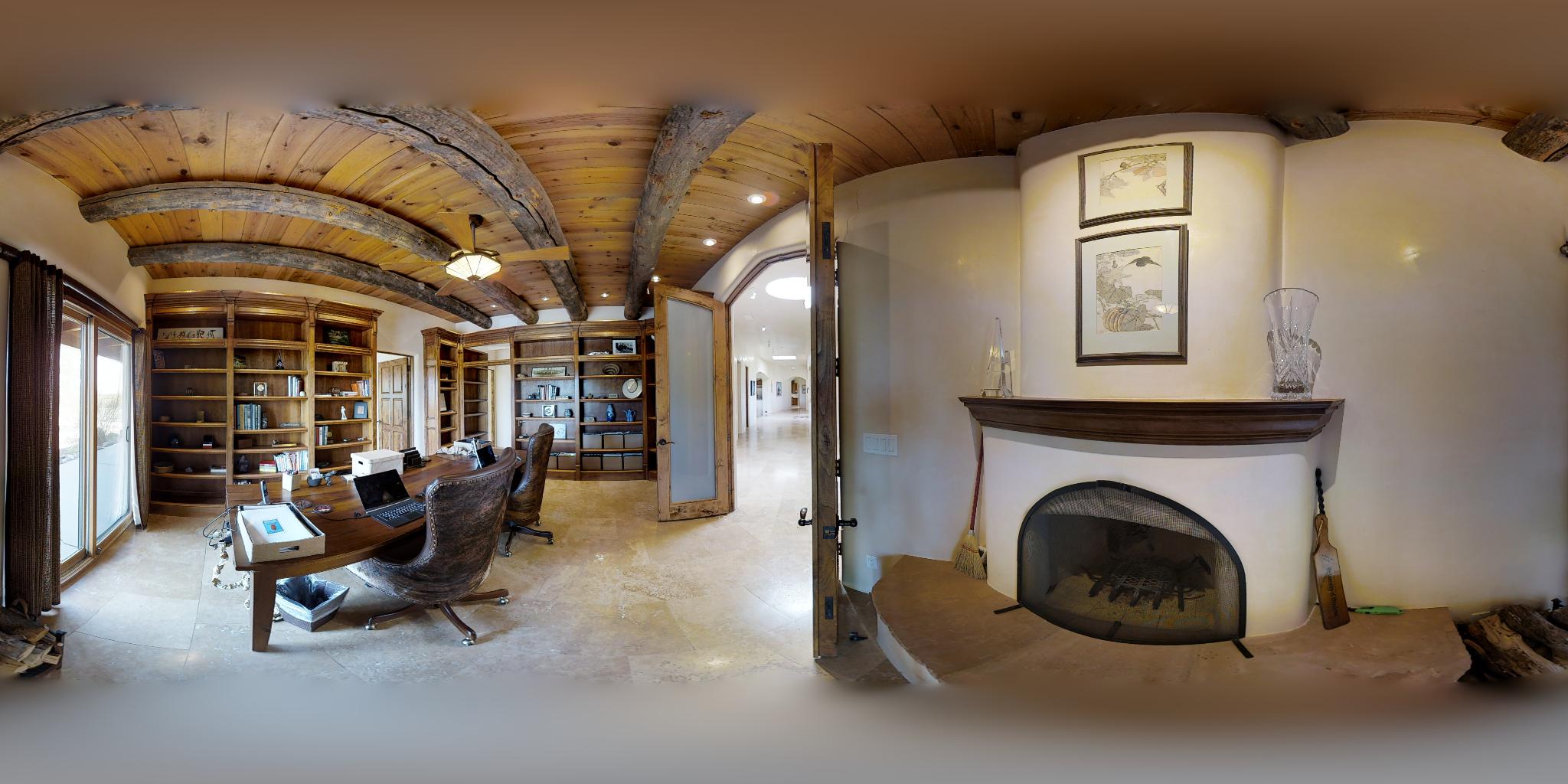}
    \label{fig:env_1_1}}
    \subfigure[]
    {\includegraphics[height=2cm]{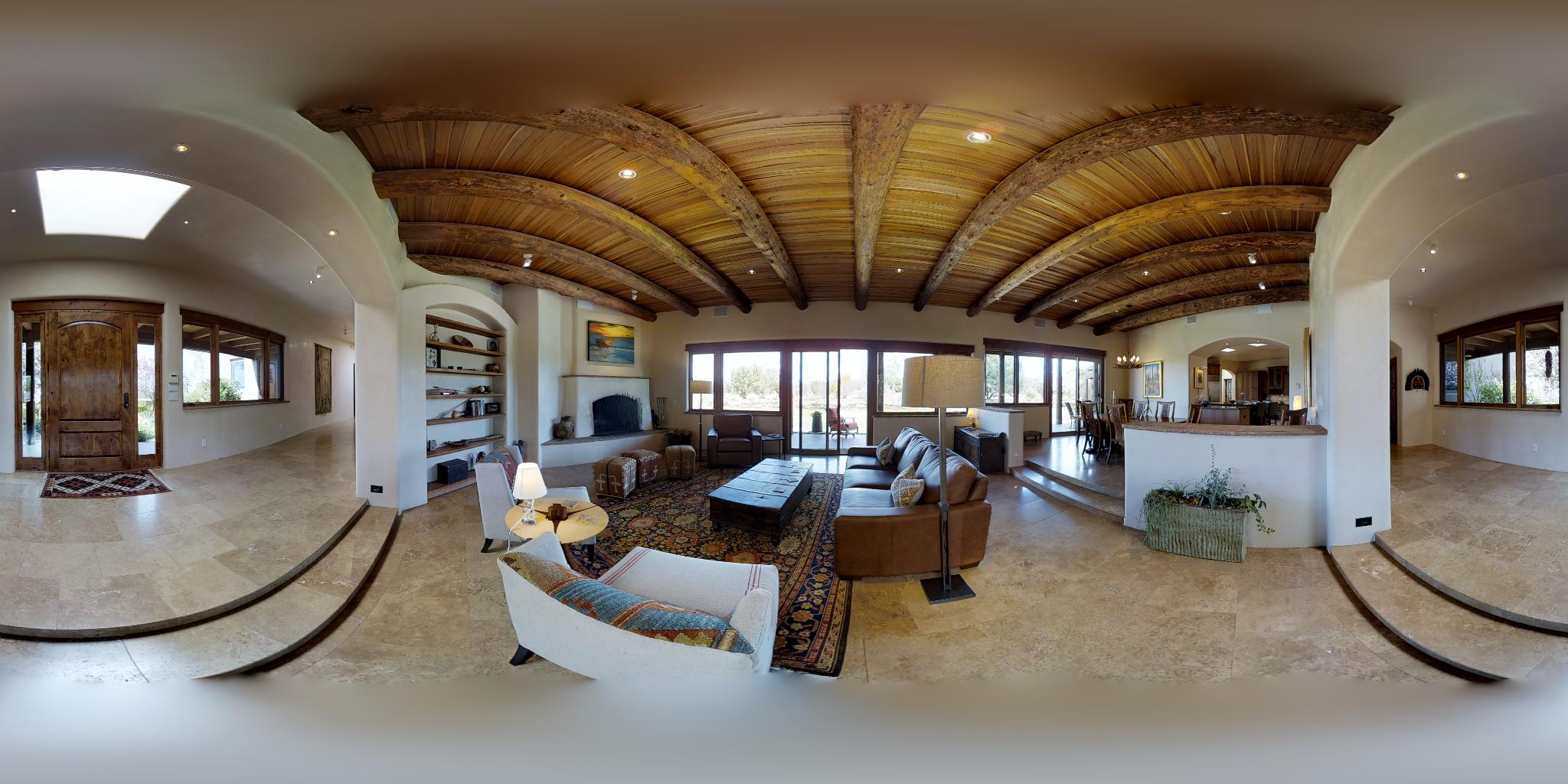}
    \label{fig:env_1_2}}
    \subfigure[]
    {\includegraphics[height=2cm]{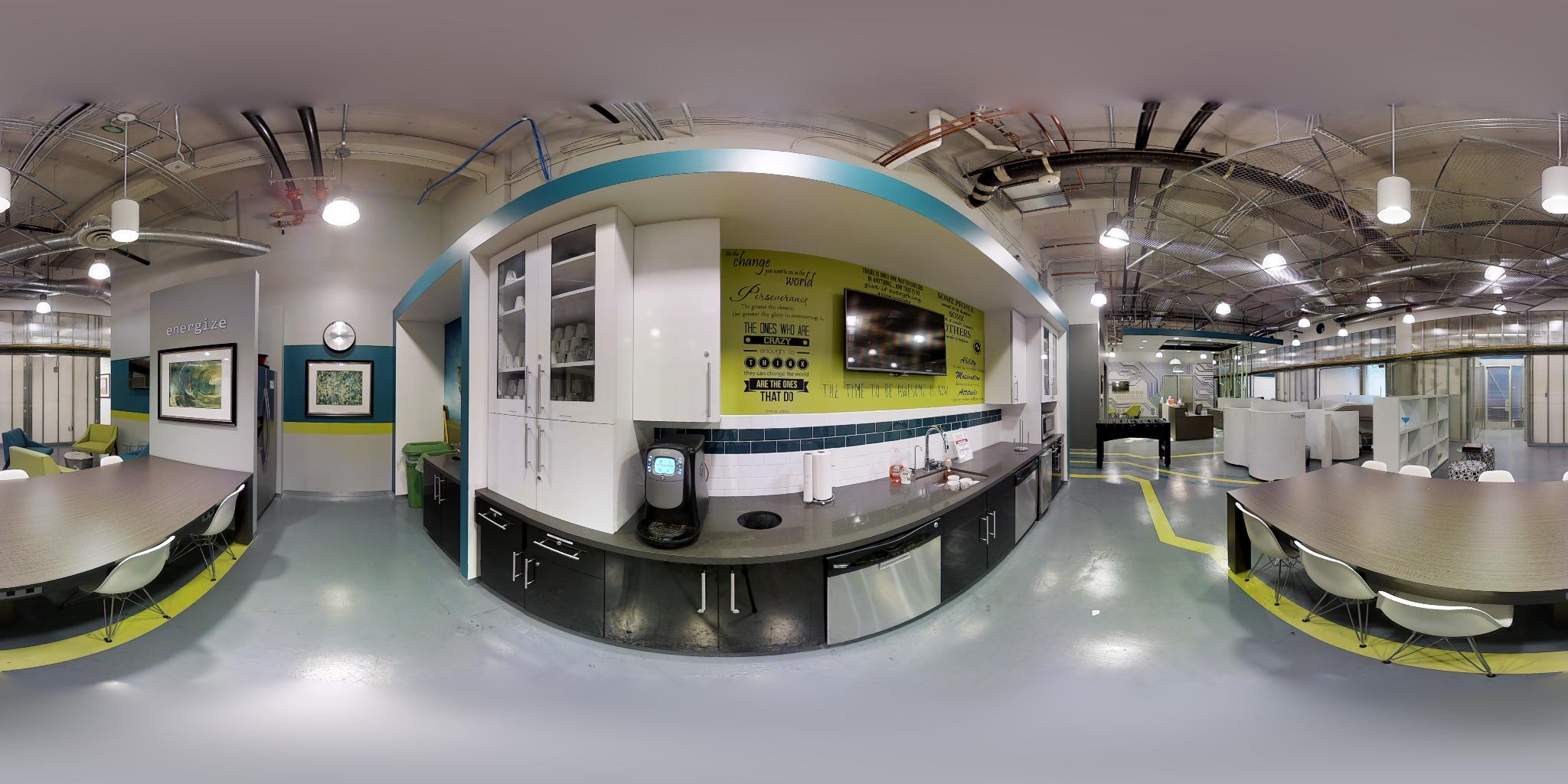}
    \label{fig:env_2_1}}
    \subfigure[]
    {\includegraphics[height=2cm]{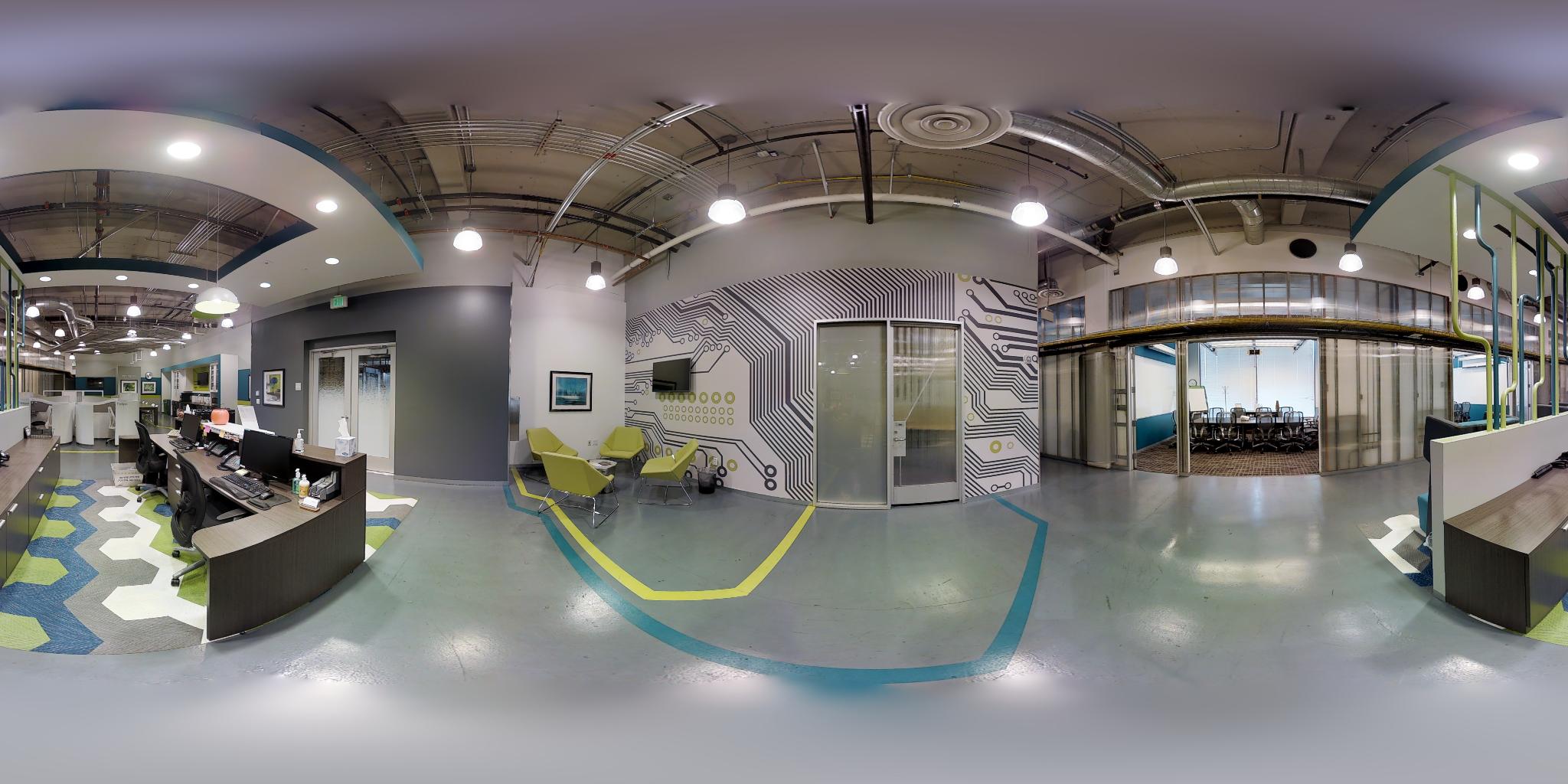}
    \label{fig:env_2_2}}
    \subfigure[]
    {\includegraphics[height=2cm]{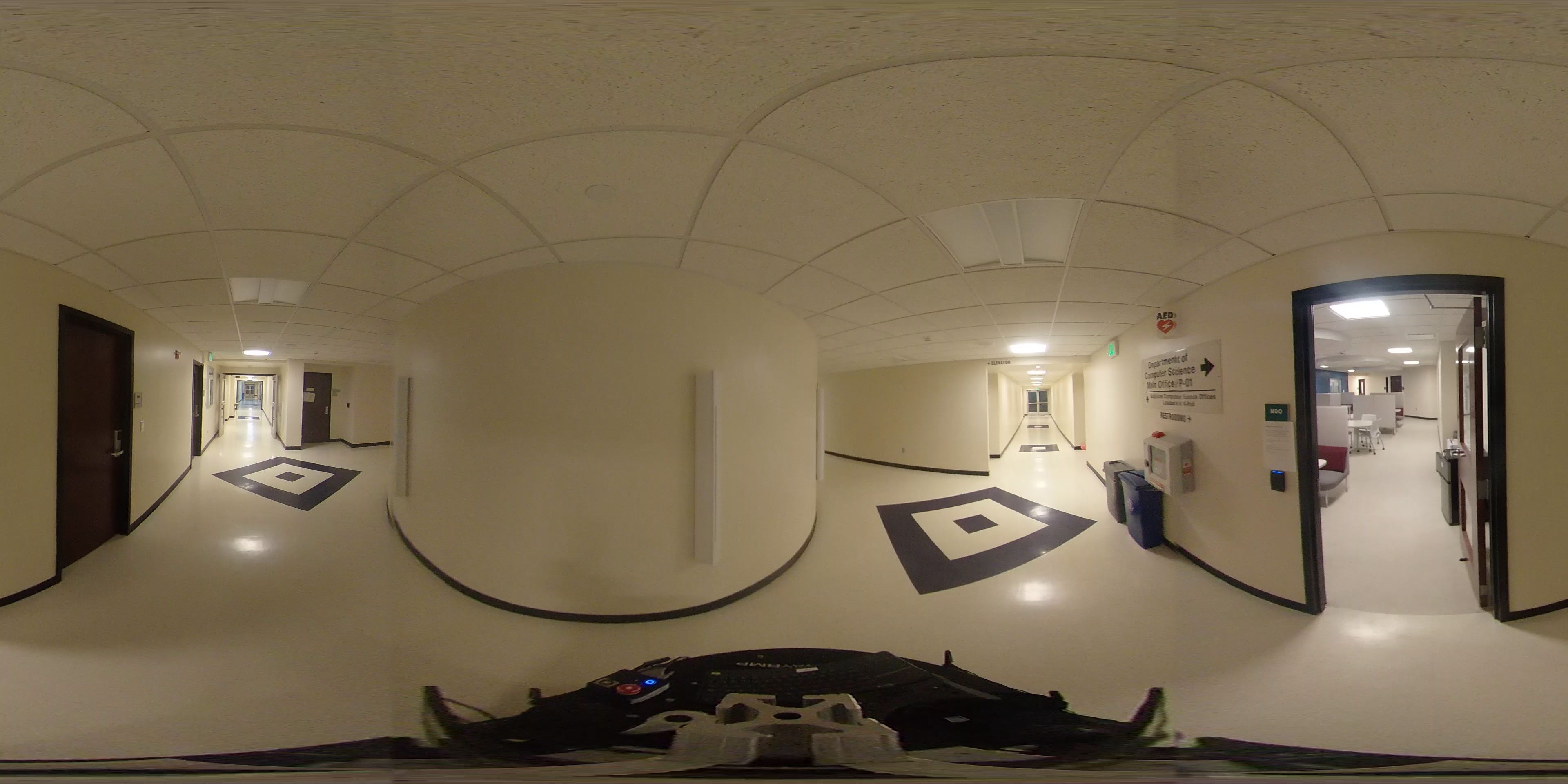}
    \label{fig:env_3_1}}    
    \subfigure[]
    {\includegraphics[height=2cm]{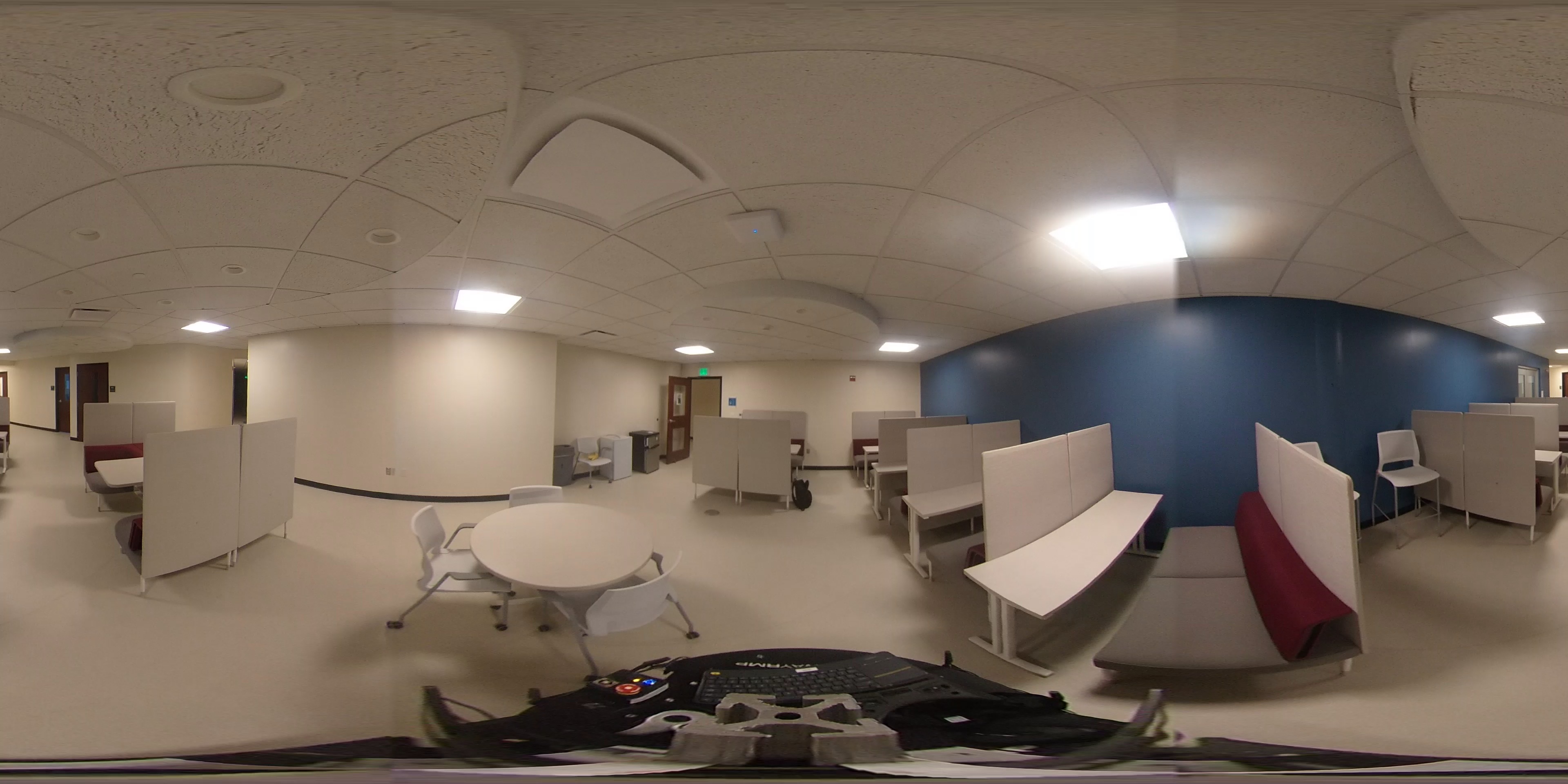}
    \label{fig:env_3_2}}
    
    \caption{(a)-(b),~Home dataset from Matterport3D; 
    (c)-(d),~Office dataset from Matterport3D; and 
    (e)-(f),~Office dataset collected using a Segway-based mobile robot platform.
    }
    \label{fig:env_figures}
    \end{center}
    \vspace{-1.5em}
\end{figure}

\subsection{Abstract Simulation} \label{sec:abstract_sim}
We use the abstract simulator to let the agent learn a strategy to guide human attention toward an object of interest in a 360$\degree$ frame.
In our abstract simulation domain, we simulate the process of GHAL360 interacting with the ``world’’ as a Markov decision process~(MDP)~\cite{puterman2014markov}.
The transition probability of the human following GHAL360’s guidance (\emph{left} or \emph{right} actions) is $0.8$ in abstract simulation. There is $0.2$ probability that the human randomly looks around in the remote environment.
The transition probability of the $left$ action and the $right$ action is 0.8 while the transition probability for the $confirm$ action is 1.0.

The reward function is in the form of $R(s,a)\rightarrow \mathbb{R}$. 
After the agent takes action $confirm$, if the value of $W_0$ (the wedge that the human is focusing on) is either $w_2$ or $w_3$, as defined in Section~\ref{sec:scene_analysis}, the agent receives a big reward of $250$. 
If the value of $W_0$ is either $w_0$ or $w_4$ after action $confirm$, the agent receives a big penalty, in the form of a reward of $-250$.
Each indication action ($left$ or $right$) produces a small cost (negative reward) of $3$ if the value of $W_0$ is $w_0$ or $w_2$.
But if the $left$ or $right$ actions result in the value of $W_0$ becoming $w_1$ or $w_3$, then the actions produces a slightly higher cost (negative) of $15$.


After executing the action ($a$), the agent moves to a next state ($s'$) and receives a reward ($r$).
The reward is based on the value of $W_{0}$ which can be one of the four values $\{w_{0},w_{1},w_{2},w_{3}\}$ as described in Section~\ref{sec:scene_analysis}, and $W_{0}$ is the wedge that the human operator is currently focusing on.
If the agent takes action $confirm$ and transitions to $s'$, where $W_{0}$ value is $w_{2}$ or $w_{3}$, the agent gets a reward as $250$.
But if the agent takes action $confirm$, and the value of $W_{0}$ is either $w_{0}$ or $w_{1}$, the agent gets a reward as $-250$.
The actions $left$ and $right$ result in a reward of $-3$ if $W_{0} = w_{0}$, whereas the reward is $-15$ if the value of $W_{0} = w_{1}$ or $w_{3}$.

We trained the agent for $15000$ episodes which resulted in the agent learning a policy for guiding human attention.
We have evaluated the learned policy in the realistic simulation.

\subsection{Realistic Simulation}

\begin{figure}[t]
    \centering
    \subfigure[Front]{\includegraphics[height=4cm]{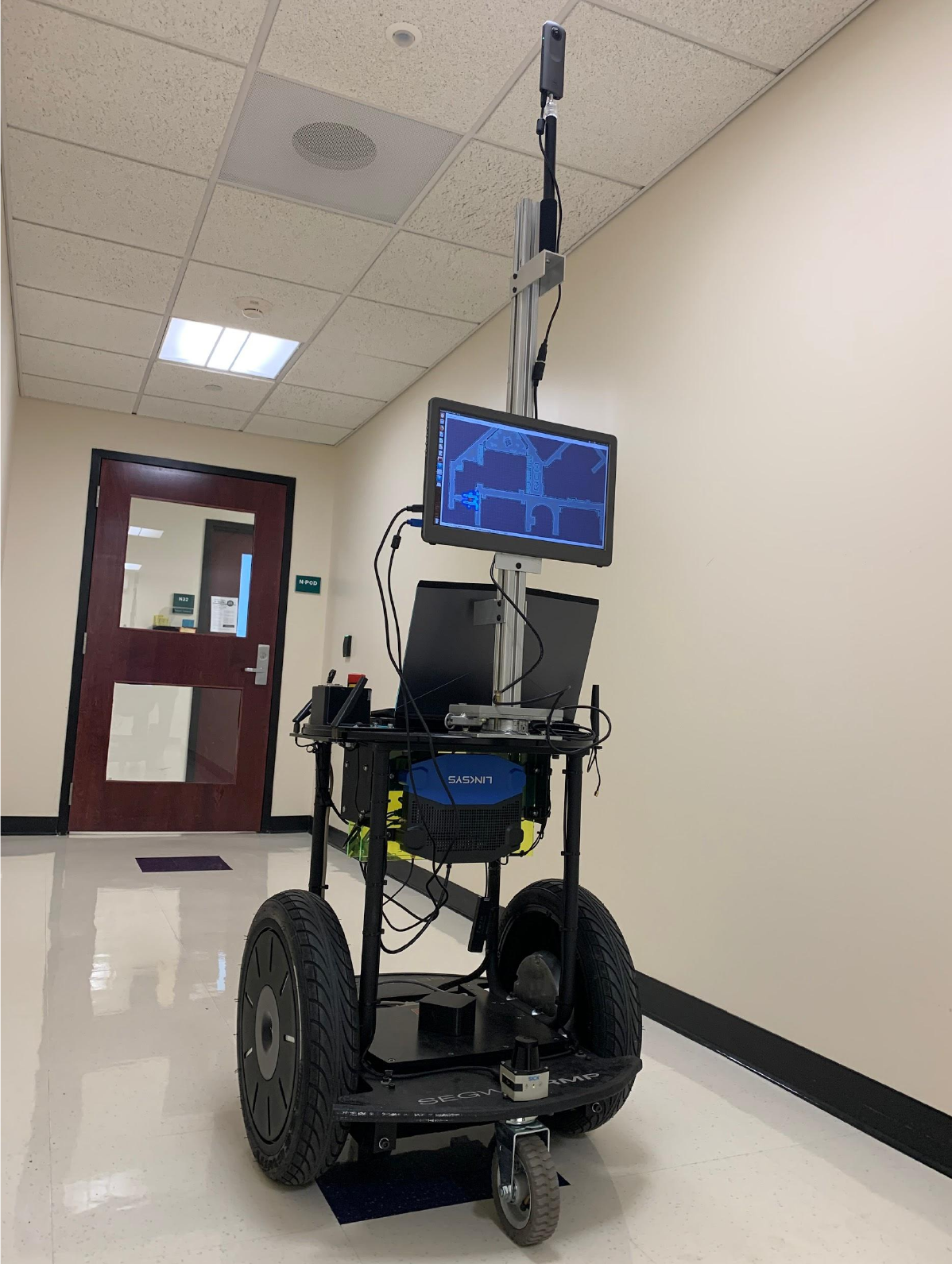}} 
    \subfigure[Back]{\includegraphics[height=4cm]{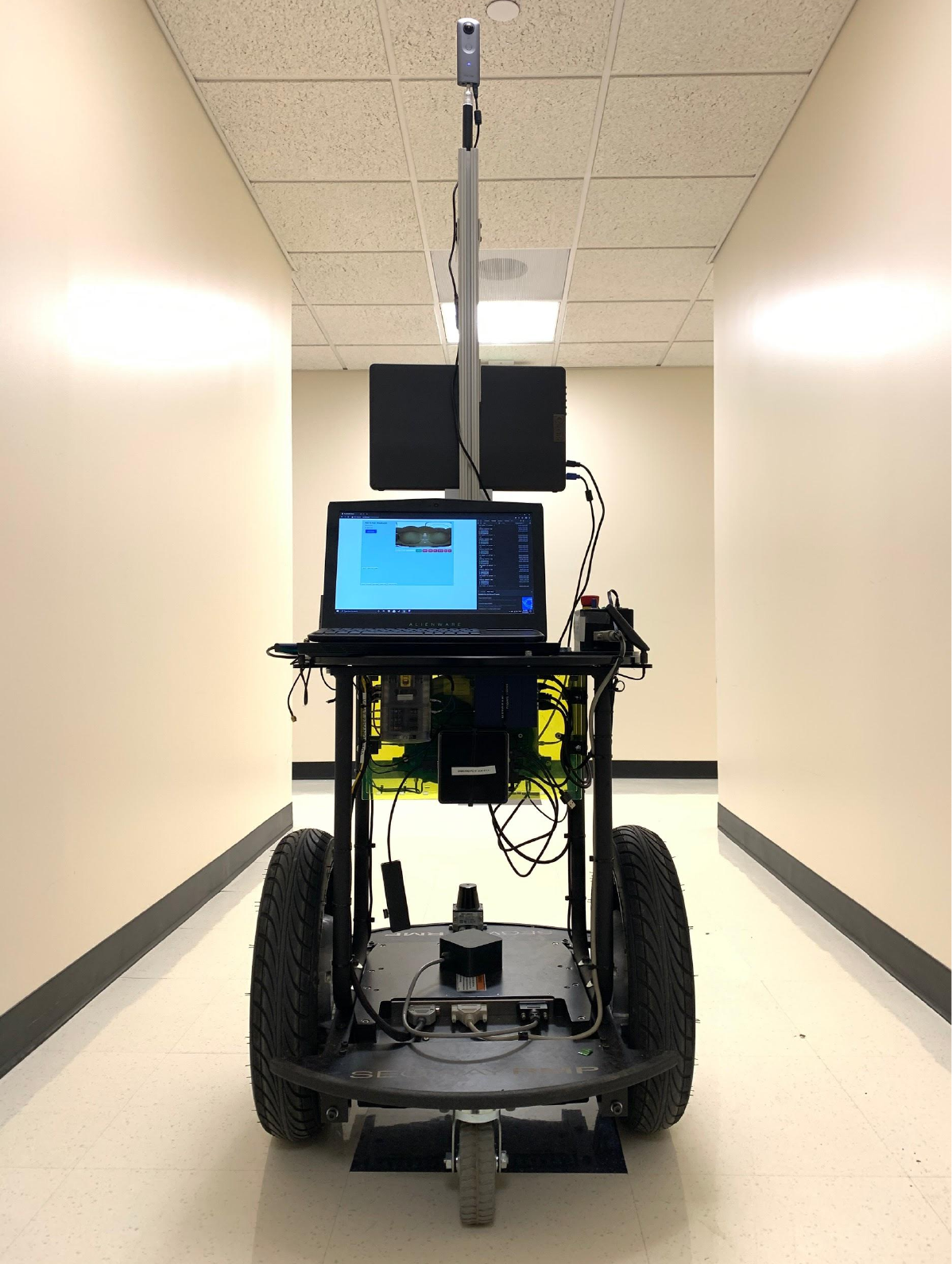}} 
    \vspace{-.5em}
    \caption{Segway-based mobile robot platform (RMP-110) with a $360\degree$ camera (Ricoh Theta V), mounted on top of the robot, and laser ranger finder (SICK TiM571 2D) that has been used for prototyping and evaluation purposes in this research.  
    }
    \vspace{-1em}
    \label{fig:robot_segway}
\end{figure}

We used the equirectangular frames from the Matterport3D datasets~(Fig.~\ref{fig:env_1_1}~-~\ref{fig:env_2_2}) as well as frames from the 360$\degree$ videos captured from the real world~(Fig.~\ref{fig:env_3_1}~-~\ref{fig:env_3_2}), and then simulated human and robot behaviors to build our realistic simulation. 
We used a Segway-based mobile robot platform to capture the $360\degree$ videos~(Fig.~\ref{fig:robot_segway}).
The robot is equipped with an on-board 360$\degree$ camera (Ricoh Theta V), and was teleoperated in a public space during the data collection process.

We model a virtual human in the realistic simulation that can also look around in the remote environment and send control signals to teleoperate the simulated robot using the interface.
Moreover, as opposed to abstract simulation, the actions of the virtual human can be visualized in the interface, for example, if the virtual human looks to the left, the viewport is changed accordingly in the interface.
The virtual human can both track the location of the simulated robot through the 2D map and perceive the 360$\degree$ view of the remote environment via the telepresence interface.
In our realistic simulation, the robot navigates in the 2D grid.
If the virtual human tries to move to an inaccessible position, the robot stays still.
Based on the control signal, the location of the robot is changed, and the interface is updated accordingly.

\subsection{Baselines vs. GHAL360}

We compared GHAL360 using a target-search scenario with different baselines from the literature.
We also conducted ablation studies to investigate how individual components of GHAL360 contribute to the system. 
The different systems used for comparison are as follows:
\begin{itemize}
    \item \textbf{MFO}: Monocular Fixed Orientation based MTR, where the remote user needs to rotate the robot to look around in the remote environment~\cite{kaplan1997internet}.
    
    \item \textbf{ADV}: All-Degree View systems which are typical MTRs equipped with a $360\degree$ FOV camera. The human can use an interface to look around in the remote environment without having to move the robot base~\cite{heshmat2018geocaching}.
    
    \item \textbf{FGS}: Fixed Guidance Strategy system which consists of a scene analysis component along with a hand-coded guidance strategy that greedily guides the human using the shortest path to the target object.
    
    \item \textbf{RLGS}: Reinforcement Learning-based Guidance Strategy system uses the policy from RL to guide the human attention and suggests an optimal path for the human operator to locate the target object. 

        
\end{itemize}

FGS and RLGS are ablations of approach, and the comparisons of FGS and RLGS with GHAL360 are ablation studies.

We used three different environments: two from the Matterport3D datasets, and one dataset collected from the real-world.
In every environment, there were six sets of start positions for the robot in each of the environments, with each set includes positions of the same distance to the target.
For each environment, we selected two target objects, and each of the target objects was specifically selected to ensure that they were unique in the environment to avoid the confusion resulting from finding a different instance of the same object at different locations.
For example, the target objects selected in environment $1$ were ``backpack'' and ``laptop'', while the target objects selected in environment $2$ were ``bottle'' and ``television''. 
The initial orientation of the robot was randomly selected, and a set of one thousand paired trials were carried out for every two-meter distance from the target object until 12 meters. 
The virtual human carried out one random action every two seconds out of ``\emph{move forward}'', ``\emph{move backward}'', ``\emph{look left}'', and ``\emph{look right}'' in the baselines (MFO and ADV).
The actions ``\emph{move forward}'', ``\emph{move backward}'' are also carried out randomly in FGS and RLGS, while the actions ``\emph{look left}'' and ``\emph{look right}'' are carried out randomly only until there are no visual indications.
In case, there are visual indications, the human follows the indications with a $0.95$ probability.
Using these actions, the simulated user could explore the remote environment.
Once the target object is in the field of view of the human operator, the trial is terminated.

\begin{figure}[tb]
    \begin{center}
    \vspace{.5em}
    \includegraphics[width=8cm]{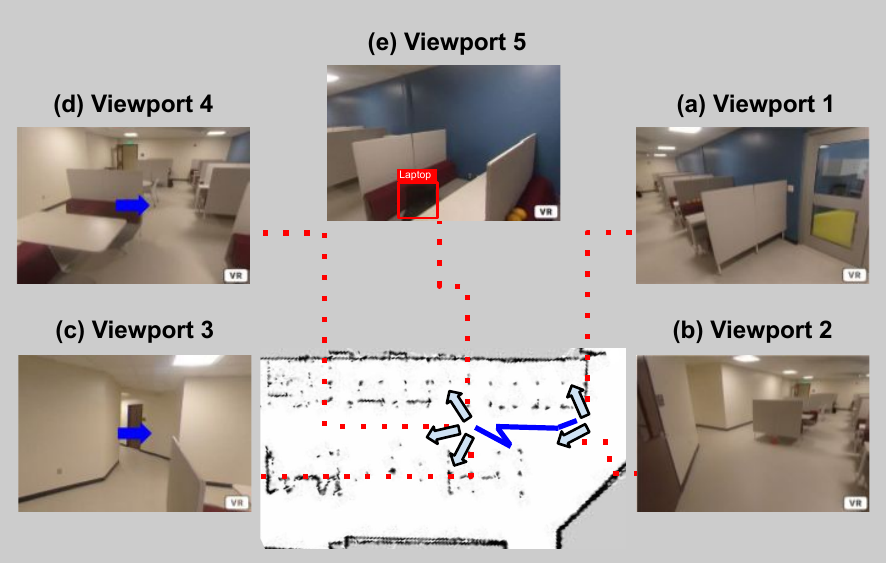}
    \end{center}
    \vspace{-.5em}
    \caption{Map showing the trajectory of the robot in the remote environment looking for a ``laptop.'' The arrows show the different human orientations along the trajectory; The images marked as (a)-(e) are the different viewports of the human at every orientation shown by the arrows.}
    \label{fig:illustrative_trial_main_fig}
\end{figure}

\subsection{Illustrative Trial}
Consider an illustrative trial where the human-MTR team needed to locate a laptop in a remote environment.
Fig.~\ref{fig:illustrative_trial_main_fig} shows the map of the remote environment with the robot’s trajectory marked in blue color.
The arrows in Fig.~\ref{fig:illustrative_trial_main_fig} indicate the different human head orientations, and the red dotted lines point to the different viewports at these orientations.
Fig.~\ref{fig:illustrative_trial_main_fig}~(a) shows the initial viewport of a virtual human.
The human started to look around at the remote environment to locate the laptop.
Fig.~\ref{fig:illustrative_trial_main_fig}~(b) shows the second viewport after the human looked left.
Then the human again looked right, and the viewport can be seen in Fig.~\ref{fig:illustrative_trial_main_fig}~(a).
The human kept looking at the same position for next two time steps.
The particle filter estimated the human's intention, and accordingly, the controller output control signals to drive the robot to the next location.

\begin{figure}[tb]
    \begin{center}
    \includegraphics[width=6.5cm]{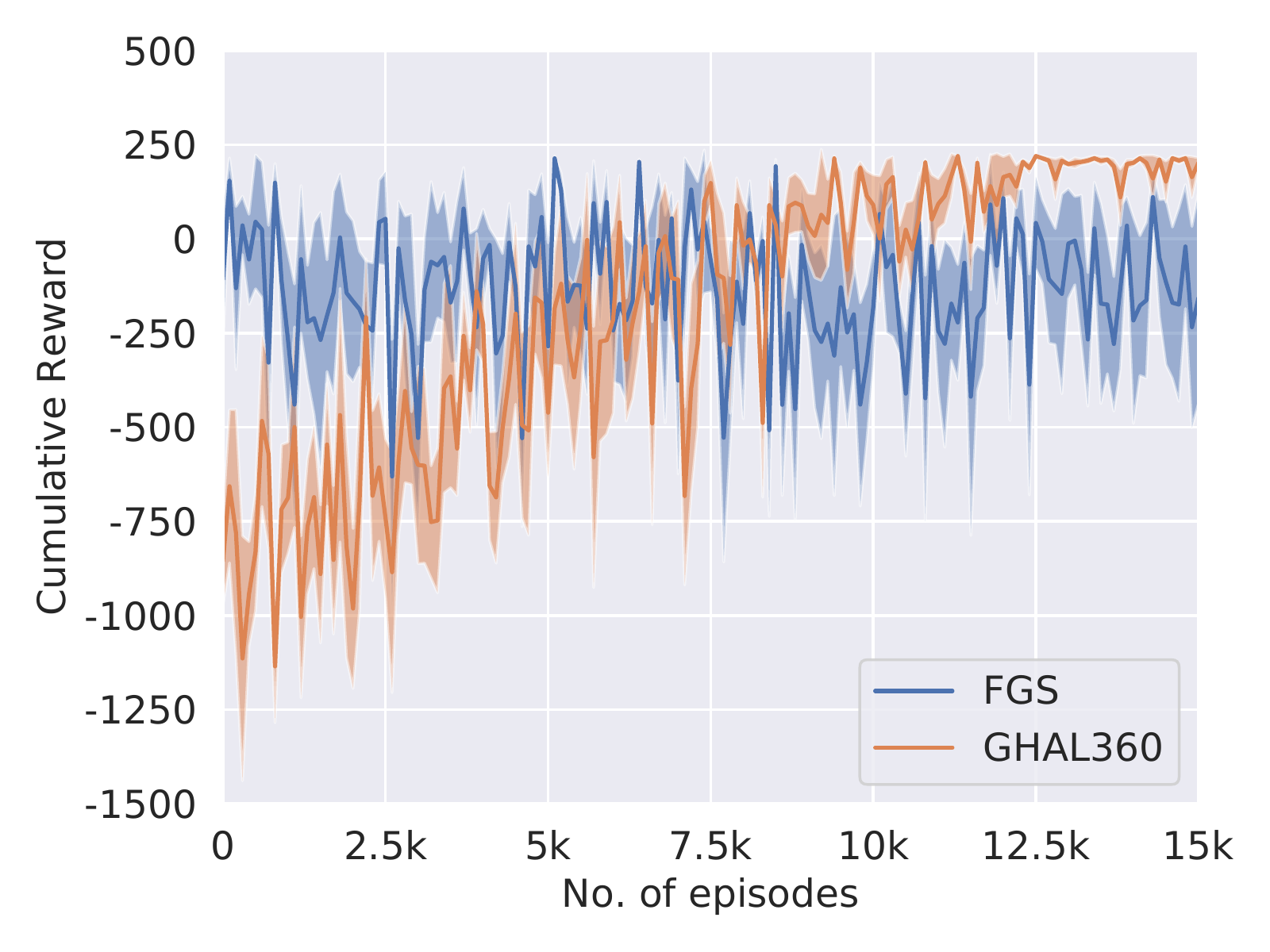}
    \end{center}
    \vspace{-1em}
    \caption{Comparison of the cumulative reward between FGS and GHAL360. Results show that the learned policy ultimately performed better than FGS (hand-coded baseline).}
    \label{fig:policy_evolution}
\end{figure}

The human continued to look around to find the laptop, and the particle filter constantly estimates the human motion intention resulting in the robot moving to different locations as shown by the blue marked trajectory in the map~(Fig.~\ref{fig:illustrative_trial_main_fig}).
Finally, the robot reached location where the target object was located.
As soon as the robot reached the location of the target object, the scene analysis component detected the laptop in the frame, and using the human head pose, GHAL360 overlayed the visual indicator over the viewport as seen in Fig.~\ref{fig:illustrative_trial_main_fig}~(d).
The indicator suggested the human to look right to find the laptop, but at first the human did not follow the guidance and looked left.
Again, the visual indicator was overlayed on the new viewport to guide the human attention~(Fig.~\ref{fig:illustrative_trial_main_fig}~(c)). 
Now, at this time step, the human looked in the guided direction, and the viewport changed back to Fig.~\ref{fig:illustrative_trial_main_fig}~(d).
Finally, the human followed the visual indicator and looked further right, and Fig.~\ref{fig:illustrative_trial_main_fig}~(e) shows the resulting viewport of the human with the laptop.
Once the human found the target object, the trial was terminated.

\begin{figure*}[t]
\begin{center}
    \subfigure[Env 1: Matterport3D (Home)]
    {\includegraphics[width=5.5cm]{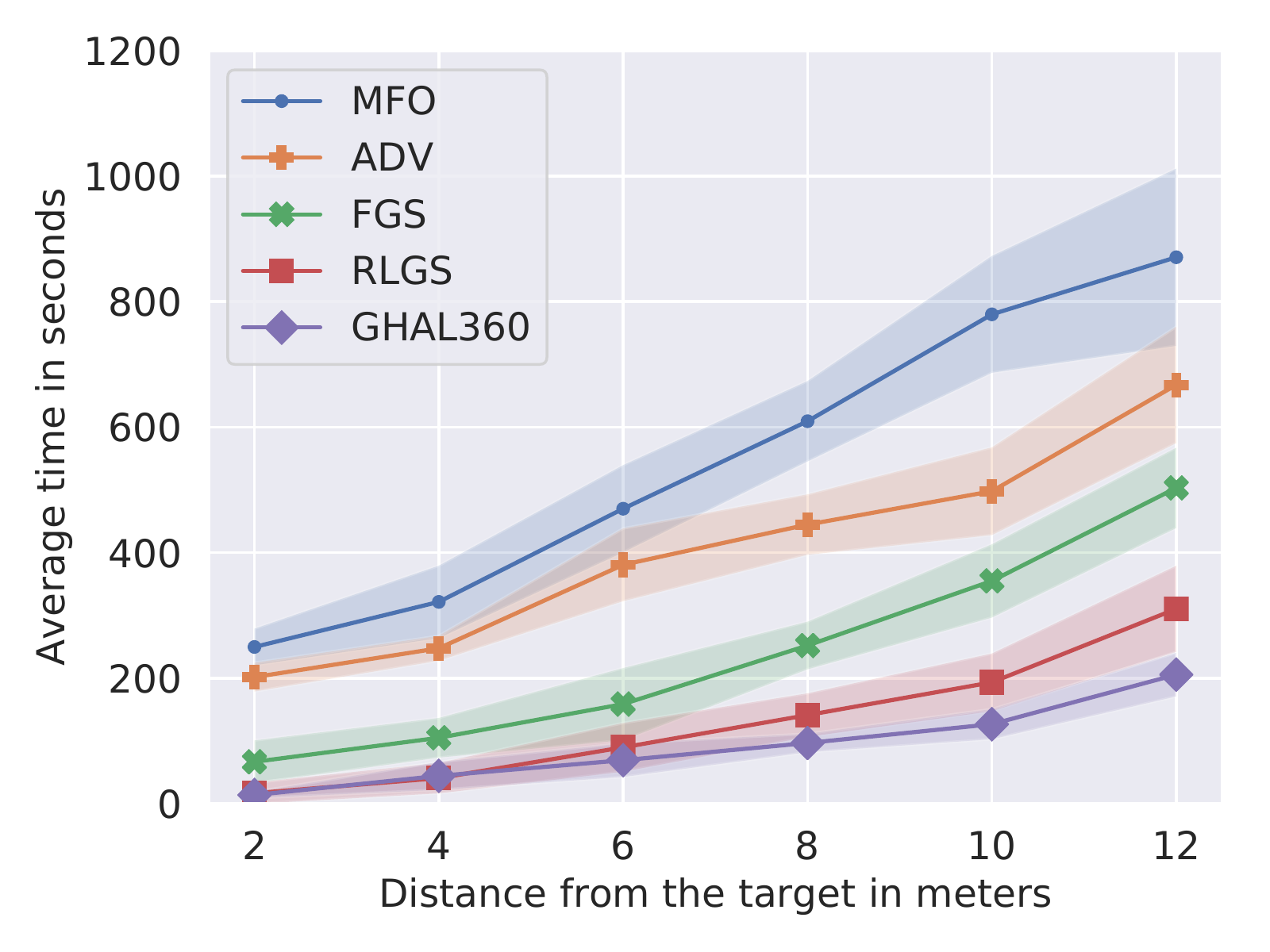}
    \label{fig:env_1_plot}}
    \subfigure[Env 2: Matterport3D (Office)]
    {\includegraphics[width=5.5cm]{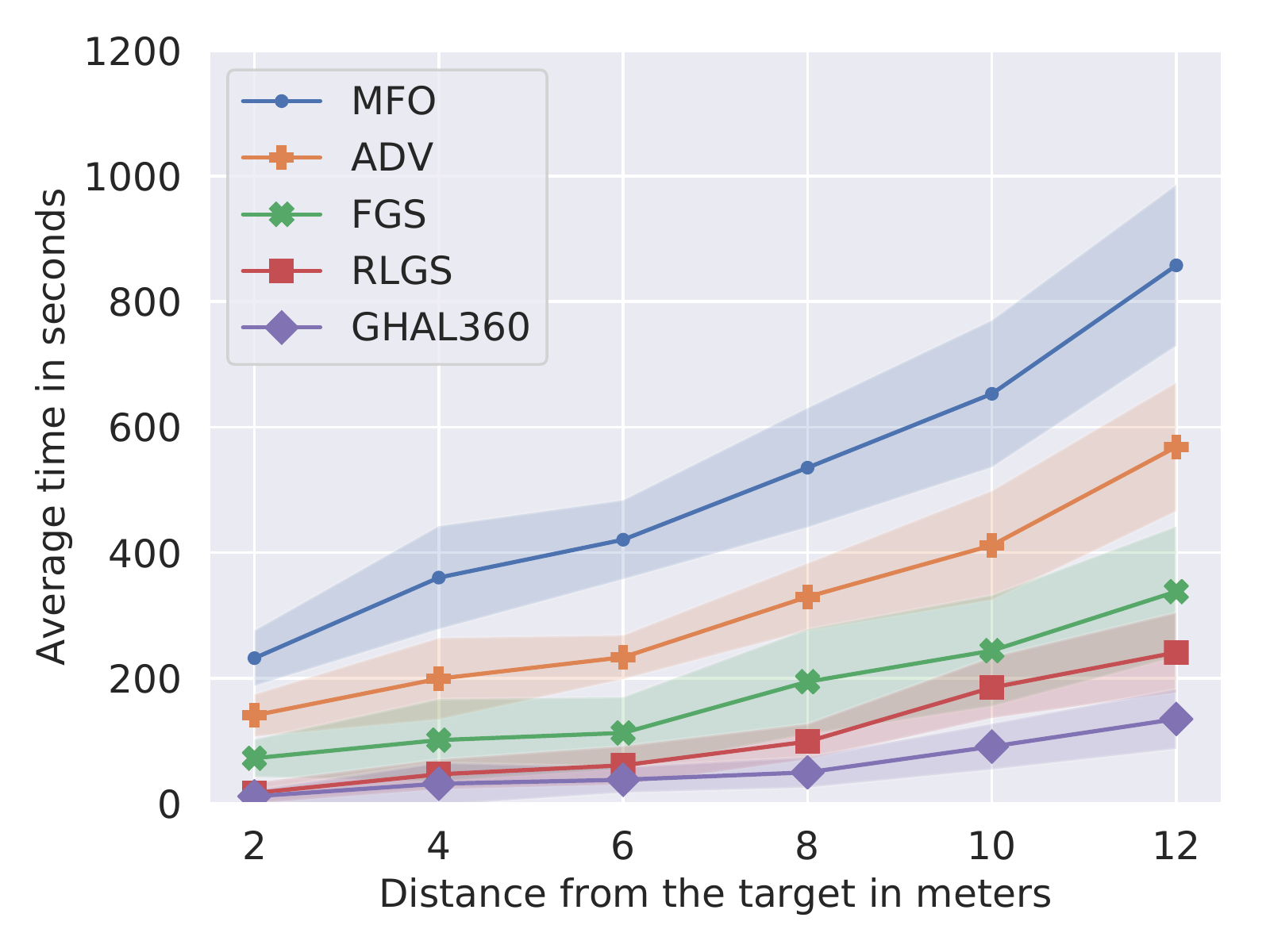}
    \label{fig:env_2_plot}}
    \subfigure[Env 3: Real Robot (Office)]
    {\includegraphics[width=5.5cm]{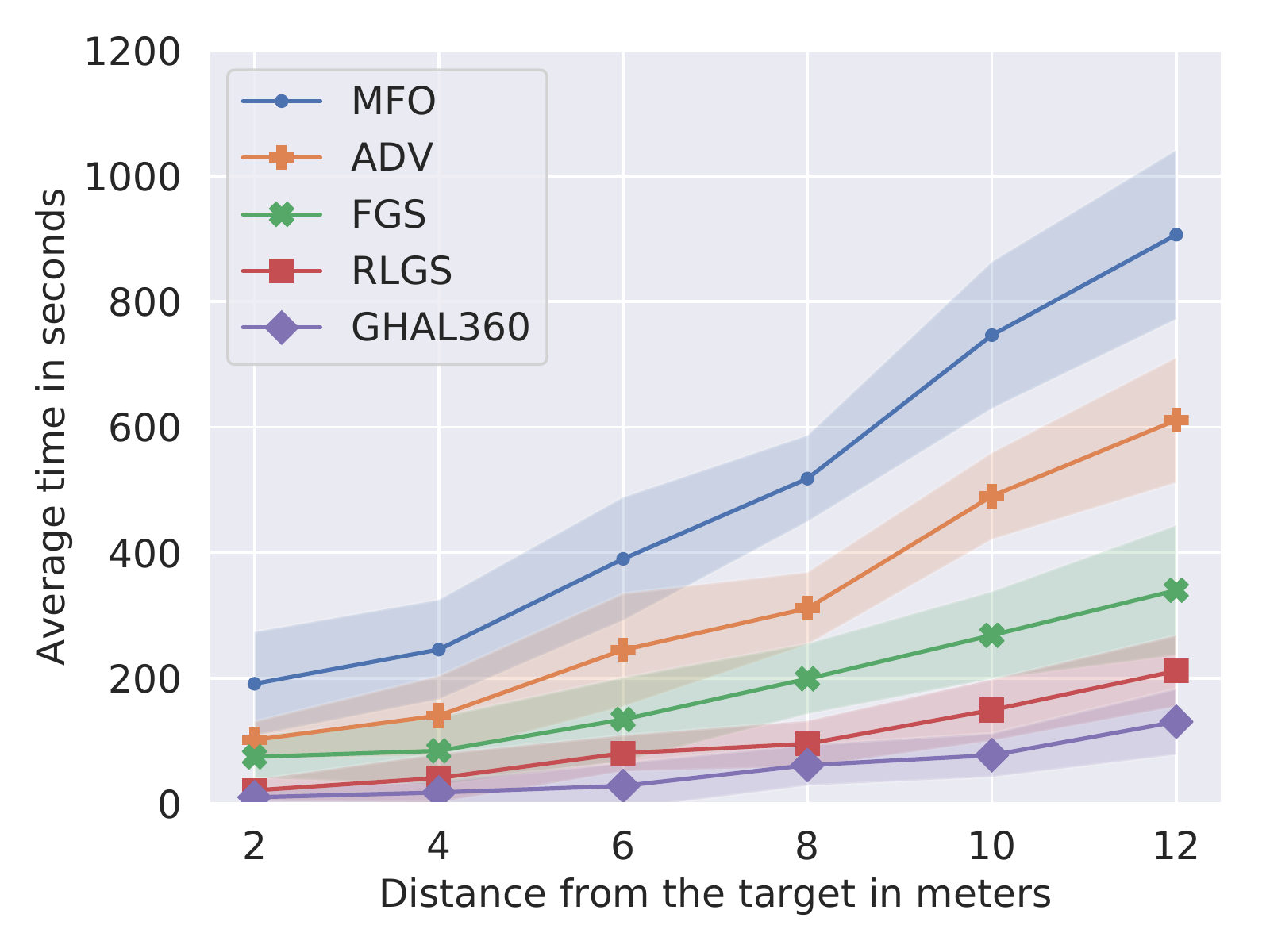}
    \label{fig:env_3_plot}}
    \caption{Comparisons are made between GHAL360 and the other systems in terms of average time (y-axis) taken to find the target object from six different distances (x-axis).
    }
    \label{fig:time_vs_distance}
    \end{center}
    \vspace{-1em}
\end{figure*}

\subsection{Results}
 
From the $15000$ training episodes, we extracted $150$ different policies after each batch of $100$ episodes, and tested them in our realistic simulation.
Fig.~\ref{fig:policy_evolution} shows the average cumulative reward and standard deviations over five runs. 
We plotted the mean cumulative reward (Fig.~\ref{fig:policy_evolution}) of FGS and GHAL360 to compare the performance of hand-coded policy with the learned policy of GHAL360.
The x-axis represents the episode number while the y-axis represents the cumulative reward.
Fig.~\ref{fig:policy_evolution} shows that in the early episodes when the agent is still exploring, the cumulative reward for GHAl360 is much lower than FGS.
But with the increasing number of episodes, it can be observed that GHAL360 outperforms FGS which represents the learned policy is superior to the hand-coded policy.

Fig.~\ref{fig:time_vs_distance} shows the overall performance of GHAL360 compared to the other systems enlisted above.
Each data point represents the average task completion time (y-axis) for six different distances (x-axis) (i.e., randomly sampled positions in each of the six distances).
One thousand trials were divided over five runs to plot each data point.
The shaded regions denote the standard deviations over five runs for each of the six distances.

From the results, it can be observed that GHAL360 outperforms all the baselines (MFO and ADV) in terms of average task completion time at all six distances in finding the target object.
Moreover, compared to FGS, GHAL360 performs better in all three environments and at six different distances denoting that the human guidance using the learned strategy outperforms the hand-coded greedy.

Furthermore, in addition to the learned guidance strategy, GHAL360 utilizes a particle filter to estimate human intention and guide the robot base toward the intended direction.
In the bottom left of Fig.~\ref{fig:env_1_plot}~-~\ref{fig:env_3_plot}, we can see that the average time of task completion is similar between RLGS and GHAL360.
This indicates that particle filter in GHAL360 is less useful when the robot starts from a position that is close to the target object’s.
When the robot is distant from the target object, we see the particle filter produces a significant improvement, as evidenced by seeing the far right of the Fig.~\ref{fig:env_1_plot}~-~\ref{fig:env_3_plot}.
This observation confirms that the particle filter also contributes to the efficiency of GHAL360.

In addition to comparing the efficiency of different systems to GHAL360, we also compared the accuracy of target search tasks as shown in Table~\ref{table:accuracy}.
It can be observed that the accuracy of GHAL360 is higher in finding all the target objects, except for the ``backpack'' object in Environment 3.
Also, we observed significant improvement in accuracy in Environment 2 for find the ``television''.
This shows that along with the improvements in efficiency, GHAL360 also provides higher accuracy in finding target objects as compared to the other systems.

\begin{table}[ht]

\scriptsize
\caption{This table shows the accuracy of different systems in target search tasks for different environments with different target objects. Bold font shows the best accuracy among all the systems, where as the bold and italic font represent the significant improvements.
  }
  \centering
\begin{center}
\setlength{\tabcolsep}{5pt}

\begin{tabular}{ |l|c|c|c|c|c| }
 \hline
 Env (target) & \multicolumn{5}{|c|}{Accuracy} \\ 
 \hline
  & MFO & ADV & FGS & RLGS & GHAL360\\ 
 \hline
 Env 1 (backpack) & .63 (.02) & .61 (.02) & .74 (.02) & .82 (.02) & \textbf{.84 (.02)}\\
 \hline
 Env 1 (laptop) & .65 (.01) & .66 (.02) & .75 (.02) & .82 (.02) & \textbf{.85 (.01)}\\
 \hline
 Env 2 (bottle) & .54 (.02) & .52 (.03) & .68 (.01) & .78 (.01) & \textbf{.81 (.02)}\\
 \hline
 Env 2 (television) & .58 (.01) & .59 (.01) & .71 (.01) & .80 (.03) & \textit{\textbf{.87 (.03)}}\\
 \hline
 Env 3 (backpack) & .55 (.02) & .57 (.02) & .68 (.03) & .82 (.03) & .82 (.02)\\
 \hline
 Env 3 (laptop) & .56 (.02) & .56 (.01) & .69 (.02) & .80 (.01) & \textbf{.83 (.02)}\\
 \hline
  Overall & .58 (.04) & .58 (.05) & .71 (.03) & .81 (.02) & \textbf{.84 (.02)}\\
 \hline

\end{tabular}
 \label{table:accuracy}
\end{center}
\vspace{-2.5em}
\end{table}

\section{Conclusions and Future Work}
In this paper, we develop a framework (GHAL360) that enables a mobile telepresence robot (MTR) to monitor the remote environment with all-degree scene analysis and actively guide the human attention to the key areas at the same time.
We conducted experiments to evaluate GHAL360 in a target search scenario.
From the results, we observed significant improvements in the efficiency of the human-MTR team in comparison to different baselines from the literature.

In the future, we would like to implement and evaluate GHAL360 on a real robot, and conduct studies with human participants for subjective analysis. 
Currently, the scene analysis component does not go beyond object detection in the current implementation of GHAL360. 
We would like to study if and how advanced scene analysis methods, such as those based on graph neural networks~\cite{scarselli2008graph}, can contribute to the system. 
Finally, our current controller suggests the robot to follow the human's intention, which can possibly be improved using a planner to actively guide the human's attention toward the direction of the target object, even when the object is not present in the immediate surroundings.



\section*{Acknowledgment}
A portion of this research has taken place at the Autonomous Intelligent Robotics (AIR) Group, SUNY
Binghamton. AIR research is supported in part by grants from the National Science Foundation (NRI-1925044), Ford Motor Company (URP Awards 2019-2021), OPPO (Faculty Research Award 2020), and SUNY Research Foundation. The work of Xiaoyang Zhang and Yao Liu is partially supported by NSF under grant CNS-1618931.

\bibliographystyle{IEEEtran}
\bibliography{IEEEabrv,ref}
\addtolength{\textheight}{-12cm}   

\end{document}